# Tensor-Tensor Product Toolbox

Canyi Lu

canyilu@gmail.com

Carnegie Mellon University

https://github.com/canyilu/tproduct

## 1 INTRODUCTION

Tensors are higher-order extensions of matrices. In recent work [2], the authors introduced the notion of the t-product, a generalization of matrix multiplication for tensors of order three. The multiplication is based on a convolution-like operation, which can be implemented efficiently using the Fast Fourier Transform (FFT). Based on t-product, there has a similar linear algebraic structure of tensors to matrices. For example, there has the tensor SVD (t-SVD) which is computable. By using some properties of FFT, we have a more efficient way for computing t-product and t-SVD in [4]. We develop a Matlab toolbox to implement several basic operations on tensors based on t-product. The toolbox is available at

https://github.com/canyilu/tproduct.

Table 1 gives the list of functions implemented in our toolbox. In section 2, we give the detailed definitions of these tensor concepts, operations and tensor factorizations. Note that we only focus on 3 way tensor in this toolbox. We will develop the same functions for $p$-way tensor [6] in the near future. We will also provide the python version soon.

Simply run `test.m` to test all the following functions.

TABLE 1: A list of functions in t-product toolbox

| Function | Description | Reference |
|---|---|---|
| `bcirc` | block circulant matrix | Equation (6) |
| `bdiag` | block diagonalization | Equation (5) |
| `tprod` | tensor-tensor product | Definition 2.1 |
| `tran` | conjugate tensor transpose | Definition 2.2 |
| `teye` | identity tensor | Definition 2.3 |
| `tinv` | tensor inverse | Definition 2.4 |
| `tsvd` | tensor singular value decomposition | Theorem 2.2 |
| `tubalrank` | tensor tubal rank | Definition 2.7 |
| `tsn` | tensor spectral norm | Definition 2.8 |
| `tnn` | tensor nuclear norm | Definition 2.9 |
| `prox_tnn` | proximal operator of tensor nuclear norm | Theorem 2.3 |
| `tqr` | tensor QR factorization | Theorem 2.4 |

**Citing.** In citing this toolbox in your papers, please use the following reference:

Canyi Lu. Tensor-Tensor Product Toolbox. Carnegie Mellon University, June 2018. https://github.com/canyilu/tproduct.
C. Lu, J. Feng, Y. Chen, W. Liu, Z. Lin, and S. Yan. Tensor robust principal component analysis with a new tensor nuclear norm. arXiv preprint arXiv:1804.03728, 2018.

The corresponding BiBTeX citations are given below:

```
@manual{lu2018tproduct,
    author       = {Lu, Canyi},
    title        = {Tensor-Tensor Product Toolbox},
    organization = {Carnegie Mellon University},
    month        = {June},
    year         = {2018},
    note         = {\url{https://github.com/canyilu/tproduct}}
}
@article{lu2018tensor,
    author  = {Lu, Canyi and Feng, Jiashi and Chen, Yudong and Liu, Wei and Lin, Zhouchen and Yan, Shuicheng},
    title   = {Tensor Robust Principal Component Analysis with A New Tensor Nuclear Norm},
    journal = {arXiv preprint arXiv:1804.03728},
    year    = {2018}
}
```

## 2 NOTATIONS AND DEFINITIONS

In this section, we introduce some notations, tensor concepts and some tensor operators implemented in our toolbox.

### 2.1 Notations

We use the same notations as [4]. We denote tensors by boldface Euler script letters, *e.g.*, $\mathcal{A}$. Matrices are denoted by boldface capital letters, *e.g.*, $\boldsymbol{A}$; vectors are denoted by boldface lowercase letters, *e.g.*, $\boldsymbol{a}$, and scalars are denoted by lowercase letters, *e.g.*, $a$. We denote $\boldsymbol{I}_n$ as the $n \times n$ sized identity matrix. The fields of real numbers and complex numbers are denoted as $\mathbb{R}$ and $\mathbb{C}$, respectively. For a 3-way tensor $\mathcal{A} \in \mathbb{C}^{n_1 \times n_2 \times n_3}$, we denote its $(i,j,k)$-th entry as $\mathcal{A}_{ijk}$ or $a_{ijk}$ and use the Matlab notation $\mathcal{A}(i,:,:)$, $\mathcal{A}(:,i,:)$ and $\mathcal{A}(:,:,i)$ to denote respectively the $i$-th horizontal, lateral and frontal slice. More often, the frontal slice $\mathcal{A}(:,:,i)$ is denoted compactly as $\boldsymbol{A}^{(i)}$. The tube is denoted as $\mathcal{A}(i,j,:)$. The inner product between $\boldsymbol{A}$ and $\boldsymbol{B}$ in $\mathbb{C}^{n_1 \times n_2}$ is defined as $\langle \boldsymbol{A}, \boldsymbol{B} \rangle = \text{Tr}(\boldsymbol{A}^* \boldsymbol{B})$, where $\boldsymbol{A}^*$ denotes the conjugate transpose of $\boldsymbol{A}$ and $\text{Tr}(\cdot)$ denotes the matrix trace. The inner product between $\mathcal{A}$ and $\mathcal{B}$ in $\mathbb{C}^{n_1 \times n_2 \times n_3}$ is defined as $\langle \mathcal{A}, \mathcal{B} \rangle = \sum_{i=1}^{n_3} \left\langle \boldsymbol{A}^{(i)}, \boldsymbol{B}^{(i)} \right\rangle$. For any $\mathcal{A} \in \mathbb{C}^{n_1 \times n_2 \times n_3}$, the complex conjugate of $\mathcal{A}$ is denoted as $\texttt{conj}(\mathcal{A})$ which takes the complex conjugate of each entry of $\mathcal{A}$. We denote $\lfloor t \rfloor$ as the nearest integer less than or equal to $t$ and $\lceil t \rceil$ as the one greater than or equal to $t$.

Some norms of vector, matrix and tensor are used. We denote the Frobenius norm as $\|\mathcal{A}\|_F = \sqrt{\sum_{ijk} |a_{ijk}|^2}$. It reduces to the vector or matrix norm if $\mathcal{A}$ is a vector or a matrix. For $\boldsymbol{v} \in \mathbb{C}^n$, the $\ell_2$-norm is $\|\boldsymbol{v}\|_2 = \sqrt{\sum_i |v_i|^2}$. The spectral norm of a matrix $\boldsymbol{A}$ is denoted as $\|\boldsymbol{A}\| = \max_i \sigma_i(\boldsymbol{A})$, where $\sigma_i(\boldsymbol{A})$'s are the singular values of $\boldsymbol{A}$. The matrix nuclear norm is $\|\boldsymbol{A}\|_* = \sum_i \sigma_i(\boldsymbol{A})$.

### 2.2 Discrete Fourier Transformation

The Discrete Fourier Transformation (DFT) plays a core role in tensor-tensor product. We give some related background knowledge and notations here. The DFT matrix $\boldsymbol{F}_n$ is of the form

$$\boldsymbol{F}_n = \begin{bmatrix} 1 & 1 & 1 & \cdots & 1 \\ 1 & \omega & \omega^2 & \cdots & \omega^{n-1} \\ \vdots & \vdots & \vdots & \ddots & \vdots \\ 1 & \omega^{n-1} & \omega^{2(n-1)} & \cdots & \omega^{(n-1)(n-1)} \end{bmatrix} \in \mathbb{C}^{n \times n},$$

where $\omega = e^{-\frac{2\pi i}{n}}$ is a primitive $n$-th root of unity in which $i = \sqrt{-1}$. Note that $\boldsymbol{F}_n / \sqrt{n}$ is an orthogonal matrix, *i.e.*,

$$\boldsymbol{F}_n^* \boldsymbol{F}_n = \boldsymbol{F}_n \boldsymbol{F}_n^* = n \boldsymbol{I}_n. \tag{1}$$

Thus $\boldsymbol{F}_n^{-1} = \boldsymbol{F}_n^* / n$. The DFT on $\boldsymbol{v} \in \mathbb{R}^n$, denoted as $\bar{\boldsymbol{v}}$, is given by

$$\bar{\boldsymbol{v}} = \boldsymbol{F}_n \boldsymbol{v} \in \mathbb{C}^n, \tag{2}$$

Computing $\bar{v}$ by using (2) costs $O(n^2)$. A more widely used method is the Fast Fourier Transform (FFT) which costs $O(n \log n)$. By using the Matlab command fft, we have $\bar{v} = \text{fft}(v)$. Denote the circulant matrix of $v$ as

$$\text{circ}(v) = \begin{bmatrix} v_1 & v_n & \cdots & v_2 \\ v_2 & v_1 & \cdots & v_3 \\ \vdots & \vdots & \ddots & \vdots \\ v_n & v_{n-1} & \cdots & v_1 \end{bmatrix} \in \mathbb{R}^{n \times n}.$$

It is known that it can be diagonalized by the DFT matrix, *i.e.*,

$$F_n \cdot \text{circ}(v) \cdot F_n^{-1} = \text{Diag}(\bar{v}), \tag{3}$$

where $\text{Diag}(\bar{v})$ denotes a diagonal matrix with its $i$-th diagonal entry as $\bar{v}_i$. The above equation implies that the columns of $F_n$ are the eigenvectors of $(\text{circ}(v))^\top$ and $\bar{v}_i$'s are the corresponding eigenvalues.

**Lemma 2.1.** *[7] Given any real vector $v \in \mathbb{R}^n$, the associated $\bar{v}$ satisfies*

$$\bar{v}_1 \in \mathbb{R} \text{ and } conj(\bar{v}_i) = \bar{v}_{n-i+2}, \ i = 2, \cdots, \left\lfloor \frac{n+1}{2} \right\rfloor. \tag{4}$$

*Conversely, for any given complex $\bar{v} \in \mathbb{C}^n$ satisfying (4), there exists a real block circulant matrix $\text{circ}(v)$ such that (3) holds.*

Now we consider the DFT on tensors. For $\mathcal{A} \in \mathbb{R}^{n_1 \times n_2 \times n_3}$, we denote $\bar{\mathcal{A}} \in \mathbb{C}^{n_1 \times n_2 \times n_3}$ as the result of DFT on $\mathcal{A}$ along the 3-rd dimension, *i.e.*, performing the DFT on all the tubes of $\mathcal{A}$. By using the Matlab command fft, we have

$$\bar{\mathcal{A}} = \text{fft}(\mathcal{A}, [\,], 3).$$

In a similar fashion, we can compute $\mathcal{A}$ from $\bar{\mathcal{A}}$ using the inverse FFT, *i.e.*,

$$\mathcal{A} = \text{ifft}(\bar{\mathcal{A}}, [\,], 3).$$

In particular, we denote $\bar{A} \in \mathbb{C}^{n_1 n_3 \times n_2 n_3}$ as a block diagonal matrix with its $i$-th block on the diagonal as the $i$-th frontal slice $\bar{A}^{(i)}$ of $\bar{\mathcal{A}}$, *i.e.*,

$$\bar{A} = \text{bdiag}(\bar{\mathcal{A}}) = \begin{bmatrix} \bar{A}^{(1)} & & & \\ & \bar{A}^{(2)} & & \\ & & \ddots & \\ & & & \bar{A}^{(n_3)} \end{bmatrix}, \tag{5}$$

where bdiag is an operator which maps the tensor $\bar{\mathcal{A}}$ to the block diagonal matrix $\bar{A}$. Also, we define the block circulant matrix $\text{bcirc}(\mathcal{A}) \in \mathbb{R}^{n_1 n_3 \times n_2 n_3}$ of $\mathcal{A}$ as

$$\text{bcirc}(\mathcal{A}) = \begin{bmatrix} A^{(1)} & A^{(n_3)} & \cdots & A^{(2)} \\ A^{(2)} & A^{(1)} & \cdots & A^{(3)} \\ \vdots & \vdots & \ddots & \vdots \\ A^{(n_3)} & A^{(n_3-1)} & \cdots & A^{(1)} \end{bmatrix}. \tag{6}$$

Just like the circulant matrix which can be diagonalized by DFT, the block circulant matrix can be block diagonalized, *i.e.*,

$$(F_{n_3} \otimes I_{n_1}) \cdot \text{bcirc}(\mathcal{A}) \cdot (F_{n_3}^{-1} \otimes I_{n_2}) = \bar{A}, \tag{7}$$

where $\otimes$ denotes the Kronecker product and $(F_{n_3} \otimes I_{n_1})/\sqrt{n_3}$ is orthogonal. By using Lemma 2.1, we have

$$\begin{cases} \bar{A}^{(1)} \in \mathbb{R}^{n_1 \times n_2}, \\ \text{conj}(\bar{A}^{(i)}) = \bar{A}^{(n_3-i+2)}, \ i = 2, \cdots, \lfloor \frac{n_3+1}{2} \rfloor. \end{cases} \tag{8}$$

Conversely, for any given $\bar{\mathcal{A}} \in \mathbb{C}^{n_1 \times n_2 \times n_3}$ satisfying (8), there exists a real tensor $\mathcal{A} \in \mathbb{R}^{n_1 \times n_2 \times n_3}$ such that (7) holds.

## 2.3 T-product and T-SVD

For $\mathcal{A} \in \mathbb{R}^{n_1 \times n_2 \times n_3}$, we define

$$\texttt{unfold}(\mathcal{A}) = \begin{bmatrix} \boldsymbol{A}^{(1)} \\ \boldsymbol{A}^{(2)} \\ \vdots \\ \boldsymbol{A}^{(n_3)} \end{bmatrix}, \ \texttt{fold}(\texttt{unfold}(\mathcal{A})) = \mathcal{A},$$

where the `unfold` operator maps $\mathcal{A}$ to a matrix of size $n_1 n_3 \times n_2$ and `fold` is its inverse operator.

**Definition 2.1.** *(T-product) [2] Let $\mathcal{A} \in \mathbb{R}^{n_1 \times n_2 \times n_3}$ and $\mathcal{B} \in \mathbb{R}^{n_2 \times l \times n_3}$. Then the t-product $\mathcal{A} * \mathcal{B}$ is defined to be a tensor of size $n_1 \times l \times n_3$,*

$$\mathcal{A} * \mathcal{B} = \texttt{fold}(\texttt{bcirc}(\mathcal{A}) \cdot \texttt{unfold}(\mathcal{B})). \tag{9}$$

T-product is equivalent to the matrix multiplication in the Fourier domain; that is, $\mathcal{C} = \mathcal{A} * \mathcal{B}$ is equivalent to $\bar{C} = \bar{A}\bar{B}$ due to (7). Indeed, $\mathcal{C} = \mathcal{A} * \mathcal{B}$ implies

$$\begin{aligned} &\texttt{unfold}(\mathcal{C}) \\ =&\texttt{bcirc}(\mathcal{A}) \cdot \texttt{unfold}(\mathcal{B}) \\ =&(\boldsymbol{F}_{n_3}^{-1} \otimes \boldsymbol{I}_{n_1}) \cdot ((\boldsymbol{F}_{n_3} \otimes \boldsymbol{I}_{n_1}) \cdot \texttt{bcirc}(\mathcal{A}) \cdot (\boldsymbol{F}_{n_3}^{-1} \otimes \boldsymbol{I}_{n_2})) \cdot ((\boldsymbol{F}_{n_3} \otimes \boldsymbol{I}_{n_2}) \cdot \texttt{unfold}(\mathcal{B})) \\ =&(\boldsymbol{F}_{n_3}^{-1} \otimes \boldsymbol{I}_{n_1}) \cdot \bar{A} \cdot \texttt{unfold}(\bar{\mathcal{B}}), \end{aligned} \tag{10}$$

where (10) uses (7). Left multiplying both sides with $(\boldsymbol{F}_{n_3} \otimes \boldsymbol{I}_{n_1})$ leads to $\texttt{unfold}(\bar{\mathcal{C}}) = \bar{A} \cdot \texttt{unfold}(\bar{\mathcal{B}})$. This is equivalent to $\bar{C} = \bar{A}\bar{B}$. This property suggests an efficient way [4] based on FFT to compute t-product instead of using (9). See Algorithm 1.

---
**Algorithm 1** Tensor-Tensor Product [4]

---
**Input:** $\mathcal{A} \in \mathbb{R}^{n_1 \times n_2 \times n_3}, \mathcal{B} \in \mathbb{R}^{n_2 \times l \times n_3}$.
**Output:** $\mathcal{C} = \mathcal{A} * \mathcal{B} \in \mathbb{R}^{n_1 \times l \times n_3}$.
1. Compute $\bar{\mathcal{A}} = \texttt{fft}(\mathcal{A}, [\,], 3)$ and $\bar{\mathcal{B}} = \texttt{fft}(\mathcal{B}, [\,], 3)$.
2. Compute each frontal slice of $\bar{\mathcal{C}}$ by

$$\bar{C}^{(i)} = \begin{cases} \bar{A}^{(i)} \bar{B}^{(i)}, & i = 1, \cdots, \lceil \frac{n_3+1}{2} \rceil, \\ \texttt{conj}(\bar{C}^{(n_3-i+2)}), & i = \lceil \frac{n_3+1}{2} \rceil + 1, \cdots, n_3. \end{cases}$$

3. Compute $\mathcal{C} = \texttt{ifft}(\bar{\mathcal{C}}, [\,], 3)$.

---

**Definition 2.2.** *(Conjugate tensor transpose) [4] The conjugate transpose of a tensor $\mathcal{A} \in \mathbb{C}^{n_1 \times n_2 \times n_3}$ is the tensor $\mathcal{A}^* \in \mathbb{C}^{n_2 \times n_1 \times n_3}$ obtained by conjugate transposing each of the frontal slices and then reversing the order of transposed frontal slices 2 through $n_3$.*

The tensor conjugate transpose extends the tensor transpose [2] for complex tensors. As an example, let $\mathcal{A} \in \mathbb{C}^{n_1 \times n_2 \times 4}$ and its frontal slices be $\boldsymbol{A}_1, \boldsymbol{A}_2, \boldsymbol{A}_3$ and $\boldsymbol{A}_4$. Then

$$\mathcal{A}^* = \texttt{fold}\left(\begin{bmatrix} \boldsymbol{A}_1^* \\ \boldsymbol{A}_4^* \\ \boldsymbol{A}_3^* \\ \boldsymbol{A}_2^* \end{bmatrix}\right).$$

**Definition 2.3.** *(Identity tensor) [2] The identity tensor $\mathcal{I} \in \mathbb{R}^{n \times n \times n_3}$ is the tensor with its first frontal slice being the $n \times n$ identity matrix, and other frontal slices being all zeros.*

It is clear that $\mathcal{A} * \mathcal{I} = \mathcal{A}$ and $\mathcal{I} * \mathcal{A} = \mathcal{A}$ given the appropriate dimensions. The tensor $\bar{\mathcal{I}} = \texttt{fft}(\mathcal{I}, [\,], 3)$ is a tensor with each frontal slice being the identity matrix.

For an $n \times n \times n_3$ tensor, an inverse exists if it satisfies the following:

**Definition 2.4.** *(Tensor inverse) [2] An $n \times n \times n_3$ tensor $\mathcal{A}$ has an inverse $\mathcal{B}$ provided that*

$$\mathcal{A} * \mathcal{B} = \mathcal{I} \text{ and } \mathcal{B} * \mathcal{A} = \mathcal{I}. \tag{11}$$

**Definition 2.5.** *(Orthogonal tensor) [2] A tensor $\mathcal{Q} \in \mathbb{R}^{n \times n \times n_3}$ is orthogonal if it satisfies $\mathcal{Q}^* * \mathcal{Q} = \mathcal{Q} * \mathcal{Q}^* = \mathcal{I}$.*

**Definition 2.6.** *(F-diagonal Tensor) [2] A tensor is called f-diagonal if each of its frontal slices is a diagonal matrix.*

**Theorem 2.2.** *(T-SVD) [4] Let $\mathcal{A} \in \mathbb{R}^{n_1 \times n_2 \times n_3}$. Then it can be factorized as*

$$\mathcal{A} = \mathcal{U} * \mathcal{S} * \mathcal{V}^*, \tag{12}$$

*where $\mathcal{U} \in \mathbb{R}^{n_1 \times n_1 \times n_3}$, $\mathcal{V} \in \mathbb{R}^{n_2 \times n_2 \times n_3}$ are orthogonal, and $\mathcal{S} \in \mathbb{R}^{n_1 \times n_2 \times n_3}$ is an f-diagonal tensor.*

Theorem 2.2 shows that any 3 way tensor can be factorized into 3 components, including 2 orthogonal tensors and an f-diagonal tensor. See Figure 1 for an intuitive illustration of the t-SVD factorization.

Note that the result of Theorem 2.2 was given first in [2] and later some other related works [1], [6]. But their proof and the way for computing $\mathcal{U}$ and $\mathcal{V}$ are not rigorous. These issues are fixed in [4]. An efficient way for computing t-SVD is given in Algorithm 2.

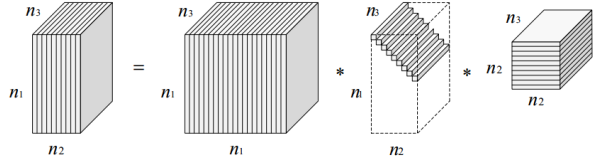

Fig. 1: An illustration of the t-SVD of an $n_1 \times n_2 \times n_3$ tensor [1].

---
**Algorithm 2** T-SVD [4]

---
**Input:** $\mathcal{A} \in \mathbb{R}^{n_1 \times n_2 \times n_3}$.
**Output:** T-SVD components $\mathcal{U}$, $\mathcal{S}$ and $\mathcal{V}$ of $\mathcal{A}$.
1. Compute $\bar{\mathcal{A}} = \texttt{fft}(\mathcal{A}, [\,], 3)$.
2. Compute each frontal slice of $\bar{\mathcal{U}}$, $\bar{\mathcal{S}}$ and $\bar{\mathcal{V}}$ from $\bar{\mathcal{A}}$ by
    **for** $i = 1, \cdots, \lceil \frac{n_3+1}{2} \rceil$ **do**
        $[\bar{U}^{(i)}, \bar{S}^{(i)}, \bar{V}^{(i)}] = \text{SVD}(\bar{A}^{(i)})$;
    **end for**
    **for** $i = \lceil \frac{n_3+1}{2} \rceil + 1, \cdots, n_3$ **do**
        $\bar{U}^{(i)} = \texttt{conj}(\bar{U}^{(n_3-i+2)})$;
        $\bar{S}^{(i)} = \bar{S}^{(n_3-i+2)}$;
        $\bar{V}^{(i)} = \texttt{conj}(\bar{V}^{(n_3-i+2)})$;
    **end for**
3. Compute $\mathcal{U} = \texttt{ifft}(\bar{\mathcal{U}}, [\,], 3)$, $\mathcal{S} = \texttt{ifft}(\bar{\mathcal{S}}, [\,], 3)$, and $\mathcal{V} = \texttt{ifft}(\bar{\mathcal{V}}, [\,], 3)$.

---

It is known that the singular values of a matrix have the decreasing order property. Let $\mathcal{A} = \mathcal{U} * \mathcal{S} * \mathcal{V}^*$ be the t-SVD of $\mathcal{A} \in \mathbb{R}^{n_1 \times n_2 \times n_3}$. The entries on the diagonal of the first frontal slice $\mathcal{S}(:,:,1)$ of $\mathcal{S}$ have the same decreasing property, *i.e.*,

$$\mathcal{S}(1,1,1) \geq \mathcal{S}(2,2,1) \geq \cdots \geq \mathcal{S}(n',n',1) \geq 0, \tag{13}$$

where $n' = \min(n_1, n_2)$. The above property holds since the inverse DFT gives

$$\mathcal{S}(i,i,1) = \frac{1}{n_3} \sum_{j=1}^{n_3} \bar{\mathcal{S}}(i,i,j), \tag{14}$$

and the entries on the diagonal of $\bar{\mathcal{S}}(:,:,j)$ are the singular values of $\bar{\mathcal{A}}(:,:,j)$. We call the entries on the diagonal of $\mathcal{S}(:,:,1)$ as the singular values of $\mathcal{A}$.

**Definition 2.7.** *(Tensor tubal rank) [4] For $\mathcal{A} \in \mathbb{R}^{n_1 \times n_2 \times n_3}$, the tensor tubal rank, denoted as $\text{rank}_t(\mathcal{A})$, is defined as the number of nonzero singular tubes of $\mathcal{S}$, where $\mathcal{S}$ is from the t-SVD of $\mathcal{A} = \mathcal{U} * \mathcal{S} * \mathcal{V}^*$. We can write*

$$\text{rank}_t(\mathcal{A}) = \#\{i, \mathcal{S}(i,i,:) \neq \mathbf{0}\}. \tag{15}$$

By using property (14), the tensor tubal rank is determined by the first frontal slice $\mathcal{S}(:,:,1)$ of $\mathcal{S}$, *i.e.*,

$$\text{rank}_t(\mathcal{A}) = \#\{i, \mathcal{S}(i,i,1) \neq 0\}. \tag{16}$$

**Definition 2.8.** *(Tensor spectral norm) [4] The tensor spectral norm of $\mathcal{A} \in \mathbb{R}^{n_1 \times n_2 \times n_3}$ is defined as $\|\mathcal{A}\| := \|bcirc(\mathcal{A})\|$.*

**Definition 2.9.** *(Tensor nuclear norm) [4] Let $\mathcal{A} = \mathcal{U} * \mathcal{S} * \mathcal{V}^*$ be the t-SVD of $\mathcal{A} \in \mathbb{R}^{n_1 \times n_2 \times n_3}$. The tensor nuclear norm of $\mathcal{A}$ is defined as*

$$\|\mathcal{A}\|_* := \langle \mathcal{S}, \mathcal{I} \rangle = \sum_{i=1}^{r} \mathcal{S}(i,i,1),$$

*where $r = \text{rank}_t(\mathcal{A})$.*

The tensor nuclear norm is the dual norm of the tensor spectral norm. By (1) and (7), both norms have the following properties

$$\|\mathcal{A}\| = \|\texttt{bcirc}(\mathcal{A})\| = \|\bar{A}\|, \tag{17}$$

$$\|\mathcal{A}\|_* = \|\texttt{bcirc}(\mathcal{A})\|_* = \|\bar{A}\|_*. \tag{18}$$

Let $g(\mathcal{Y}) = \|\mathcal{Y}\|_*$. Then the proximal operator $\text{Prox}_g^\tau(\mathcal{Y})$ of TNN is defined as

$$\text{Prox}_g^\tau(\mathcal{Y}) = \arg \min_{\mathcal{X} \in \mathbb{R}^{n_1 \times n_2 \times n_3}} \tau \|\mathcal{X}\|_* + \frac{1}{2}\|\mathcal{X} - \mathcal{Y}\|_F^2. \tag{19}$$

The computation of the proximal operator of TNN is a key subproblem in low-rank optimization [4], [5], [3]. It has a closed-form solution.

**Theorem 2.3.** *[4] For any $\tau > 0$ and $\mathcal{Y} \in \mathbb{R}^{n_1 \times n_2 \times n_3}$, the proximal operator $\text{Prox}_g^\tau(\mathcal{Y})$ (19) of TNN obeys*

$$Prox_g^\tau(\mathcal{Y}) = \mathcal{U} * \mathcal{S}_\tau * \mathcal{V}^*, \tag{20}$$

*where*

$$\mathcal{S}_\tau = \texttt{ifft}((\bar{\mathcal{S}} - \tau)_+, [\,], 3), \tag{21}$$

*and $t_+$ denotes the positive part of t, i.e., $t_+ = \max(t, 0)$.*

See Algorithm 3 for an efficient way for computing $\text{Prox}_g^\tau(\mathcal{Y})$.

---

**Algorithm 3** Tensor Singular Value Thresholding (t-SVT) [4]

---

**Input:** $\mathcal{Y} \in \mathbb{R}^{n_1 \times n_2 \times n_3}$, $\tau > 0$.
**Output:** $\text{Prox}_g^\tau(\mathcal{Y})$ as defined in (20).
1. Compute $\bar{\mathcal{Y}} = \texttt{fft}(\mathcal{Y}, [\,], 3)$.
2. Perform matrix SVT on each frontal slice of $\bar{\mathcal{Y}}$ by
    **for** $i = 1, \cdots, \lceil \frac{n_3+1}{2} \rceil$ **do**
        $[\boldsymbol{U}, \boldsymbol{S}, \boldsymbol{V}] = \text{SVD}(\bar{\boldsymbol{Y}}^{(i)})$;
        $\bar{\boldsymbol{W}}^{(i)} = \boldsymbol{U} \cdot (\boldsymbol{S} - \tau)_+ \cdot \boldsymbol{V}^*$;
    **end for**
    **for** $i = \lceil \frac{n_3+1}{2} \rceil + 1, \cdots, n_3$ **do**
        $\bar{\boldsymbol{W}}^{(i)} = \texttt{conj}(\bar{\boldsymbol{W}}^{(n_3-i+2)})$;
    **end for**
3. Compute $\text{Prox}_g^\tau(\mathcal{Y}) = \texttt{ifft}(\bar{\mathcal{W}}, [\,], 3)$.

---

Beyond t-SVD, we also have the tensor QR factorization.

**Theorem 2.4.** *(T-QR) [2] Let $\mathcal{A} \in \mathbb{R}^{n_1 \times n_2 \times n_3}$. Then it can be factorized as*

$$\mathcal{A} = \mathcal{Q} * \mathcal{R}, \tag{22}$$

*where $\mathcal{U} \in \mathbb{R}^{n_1 \times n_1 \times n_3}$ is orthogonal, and $\mathcal{R} \in \mathbb{R}^{n_1 \times n_2 \times n_3}$ is an f-upper triangular tensor (each frontal slice is an upper triangular matrix).*